\documentclass[conference]{IEEEtran}
\usepackage{hyperref}
\hypersetup{
  colorlinks   = true,    
  urlcolor     = blue,    
  linkcolor    = red,    
  citecolor    = red      
}
\usepackage{amsmath}
\usepackage{graphicx}
\usepackage{todonotes}
\usepackage{comment}

\usepackage{color}
\usepackage{algorithm,algorithmic}
\usepackage{url}

\title{Generating the support with extreme value losses}
\author{Nicholas Guttenberg$^{1,2}$ \\
\mbox{}\\
$^1$ Earth-life Science Institute, Tokyo, Japan \\
$^2$ Araya Inc, Tokyo, Japan}

\begin{document}

\maketitle
\begin{abstract}
When optimizing against the mean loss over a distribution of predictions in the context of a regression task, then even if there is a distribution of targets the optimal prediction distribution is always a delta function at a single value. Methods of constructing generative models need to overcome this tendency. We consider a simple method of summarizing the prediction error, such that the optimal strategy corresponds to outputting a distribution of predictions with a support that matches the support of the distribution of targets --- optimizing against the minimal value of the loss given a set of samples from the prediction distribution, rather than the mean. We show that models trained against this loss learn to capture the support of the target distribution and, when combined with an auxiliary classifier-like prediction task, can be projected via rejection sampling to reproduce the full distribution of targets. The resulting method works well compared to other generative modeling approaches particularly in low dimensional spaces with highly non-trivial distributions, due to mode collapse solutions being globally suboptimal with respect to the extreme value loss. However, the method is less suited to high-dimensional spaces such as images due to the scaling of the number of samples needed in order to accurately estimate the extreme value loss when the dimension of the data manifold becomes large.
\end{abstract}

\section{Introduction}

Often, rather than just learning to approximate a function mapping from input to output, we want to learn to generate samples randomly according to some distribution which is specified implicitly via data (most commonly data drawn from that distribution directly, or drawn from an associated conditional distribution). For example, one might want to generate the distribution of images which could be classified as a cat, or the distribution of actions which would all lead to an agent's behavior satisfying some constraint, or abstractly the distribution of predictions which would be consistent with a given set of observations. Having access to the distribution of outcomes can help detect anomalies, maintain uncertainty information throughout a complex model flow, or permit optimization over implicit options associated with the residual entropy of the generated distributions. Sometimes, rather than generating the distribution of data itself, we would like to generate a related distribution instead. If for example we were training an agent on a new environment, and some outcomes were extremely common in the data we had accumulated so far, we could optimize the agent's exploration by trying to shift the distribution of actions attempted in order to flatten the distribution of outcomes. That way, if there are rare patterns of actions which can achieve very different outcomes, we could enhance the sampling of those possibilities. 

We can generally divide methods of producing diverse samples into two categories. One is to build a model that outputs the parameters of some distribution which we can sample from, such as a Gaussian distribution or explicit probabilities over a finite set of discrete outcomes. The other is to transform a distribution of inputs (such as by adding noise variables) into a distribution of outputs. The first method is used by things such as auto-regressive models and mixture density networks\cite{bishop1994mixture}. Meanwhile, the second method is used as the sole source of entropy by generative adversarial networks\cite{goodfellow2014generative} and invertible flows\cite{rezende2015variational}, and is combined with the first method in the case of variational auto-encoders (VAEs)\cite{kingma2013auto}. 

The problem with passing noise through a network in order to produce a distribution of outcomes is that when optimizing over the expectation value of a scalar loss function, the optimal solution is always a single point, not a distribution. As such, there is a pressure on such networks to collapse towards a weak estimate (say, of the mean of the target value) rather than to retain the entropy of the inputs. In invertible flows, this pressure is prevented from functioning via the application of the invertibility constraint --- the entropy of the input cannot be decreased, and so the output remains a distribution. In VAEs, explicit regularization is used to reward the network for keeping the noise around. In GANs, the only thing maintaining the pressure is the dynamics of training, and so there are often issues with mode-collapse as the entropy of the output distribution irreversibly decreases.

Recent work on time-agnostic prediction\cite{jayaraman2018time} makes use of the technique of optimizing against the minimum of the loss over a set of outputs rather than the expectation. In that paper, the purpose is to give the network a degree of freedom in deciding which of a sequence of frames to 'try' to predict. By using the minimum, so long as at least one element of the sequence is predicted well, the network is not penalized for errors in any of the other elements and as such is free to allow those elements to vary arbitrarily. There is a similarity between this idea and the idea of diversity preservation in genetic algorithms via the use of a minimal criterion for reproduction rather than using a fitness function\cite{lehman2010revising} --- there as well, by constructing an objective function with large flat areas, those flat areas are free to be utilized in order to hedge against uncertainties.

Based on this comparison, we consider what would happen if rather than using the minimum loss over a concrete variable such as time, we instead had a stochastic network and took the minimum loss over many samples from the distribution of the network's output. We can then think of a batch of such samples, where we have simply replaced the batch-wise expectation of the loss with the batch-wise minimum over the loss. We show that in the context of regression problems training against the minimum regression loss between output and target causes the optimum output of the model to converge to the support of the distribution of possible outputs rather than to a single optimal point estimate as the number of samples becomes asymptotically large. Furthermore, it is possible to extend this to a generative model matching the distribution by simultaneously predicting the probability that each output will be selected as the minimum, and resampling according to that predicted probability.

Code for our numerical experiments is available at \url{https://github.com/ngutten/ExtremeValueLoss}.

\section{Related work}

\textbf{Mixture density networks (MDNs)}: Mixture density networks\cite{bishop1994mixture} directly parameterize the output distribution with (generally) a sum of Gaussians, and then train the network to maximize the likelihood of the empirical distribution under the model. When the dimension of the target space is low, this is an effective way to capture complex multi-modal structure in the joint distribution. However, the covariance matrices to specify each Gaussian grow quadratically in the dimension of the target space, and the number of Gaussians needed to fit a curved manifold structure in high dimension may likewise grow quickly. MDNs were used in SketchRNN\cite{ha2017neural} to model the continuation of lines in drawing figures and kanji.

\textbf{Auto-regressive models (ARs)}: One method for generating samples from a high-dimensional distribution is to factorize the joint distribution into a product of one-dimensional conditional probabilities: $p(x,y,z,...) = p(x)p(y|x)p(z|x,y)...$. These one-dimensional distributions can each be learned separately by discretizing the possibilities and minimizing the categorical cross-entropy or KL divergence. Alternately, if the variables $x,y,z,...$ have some structural relationship, such as words in a sentence or pixels in an image, then weight sharing and causal masking can be used to learn a single model which is able to generate all such distributions. Sampling from the distribution is done by sampling from each one-dimensional distribution in turn, conditioning on the previous samples. Alternately, methods such as beam-search can be used to find maximum likelihood outcomes or otherwise adjust the 'temperature' of the generator. Methods which use this approach include CharRNN\cite{sutskever2018generating}, PixelRNN\cite{oord2016pixel}, PixelCNN\cite{van2016conditional, salimans2017pixelcnn++}, WaveNet\cite{van2016wavenet}, and Transformer networks\cite{vaswani2017attention}.

A strong advantage of auto-regressive models is that they permit explicit calculation of a differentiable probability density around a sample, meaning that they can be used for anomaly detection, importance sampling, or other applications which require that information. The primary disadvantages of auto-regressive models are firstly that they must generally be sampled from sequentially (which makes them relatively slow compared to other methods when it comes to generation, in the worst case being effectively $O(N^2)$ in the data dimension), and secondly that when the variables in question do not share some equivalency such as in the case of structured data, a separate model must be trained for each variable. Auto-regressive models also do not possess a readily manipulable latent space which maps to the output distribution, so they cannot be readily used for things such as embedding or interpolation between samples. Additionally, the actual sampling procedure is not itself differentiable, so using auto-regressive models as a forward model of dynamics for subsequent optimization via gradient descent requires a reparameterization step which can introduce additional variance.

\textbf{Boltzmann machines}: For binary data, Boltzmann machines\cite{ackley1985learning} and Restricted Boltzmann machines (RBMs)\cite{hinton2002training} can directly learn an energy function $H$ over the space, such that the Boltzmann distribution over the space $p(x) = \frac{1}{Z} e^{-\beta H(x)}$ is matched to the empirical distribution. 

Boltzmann machines provide a direct estimation of the likelihood of samples, which can be useful for other applications. In the form of the hidden nodes, Boltzmann machines can learn a discrete latent space representing distributions over samples. However, Boltzmann machines are constrained in that the energy function must be constrained such that marginalization over the latent degrees of freedom can be done analytically, otherwise extensive Monte-Carlo simulation is needed for every update. This is generally a quadratic model with (in the case of the RBM) no within-layer links. This can make them tricky to generalize to new domains.

\textbf{Invertible flows}: This set of methods belongs to the class of techniques which learn a map between an easily generated distribution (such as a high dimensional Gaussian) and a target distribution. The idea is that, if the learned map is invertible, then it is possible to take a point in the target space, find its pre-image in the original distribution, and (in combination with the Jacobian of the map) thereby calculate its probability density. If the probability density of samples can be calculated, then the map can be optimized via gradient descent to minimize the KL divergence between the empirical distribution and generated distribution. In order to do this, the map itself cannot be an arbitrary neural network, but must be constructed out of layers for which the existence of an inverse can be guaranteed (and for which the inverse is easily constructed). The general structure of this is that alternating subsets of the dimensions are transformed according to an invertible matrix constructed from an arbitrary function of the residual dimensions, so that with respect to inversion all that is needed is to invert (what can be treated as) a constant matrix at each step. Examples of work which develops these methods are Normalizing Flows\cite{rezende2015variational}, RealNVP\cite{dinh2016density}, and Glow\cite{kingma2018glow}.

Like auto-regressive models, invertible flows provide a probability density around generated samples. Furthermore, the latent space associated with the source distribution can be used to embed points and disentangle independently varying factors from within the full distribution. The primary downsides are that these models tend to require a much greater depth than other methods (generally hundreds of layers), and due to the invertibility constraint they must maintain a constant width equal to the output dimension. As a result, though sample quality is good, methods such as Glow (which used 576 convolutional layers for the highest resolution experiments) can be relatively resource intensive to train.

\textbf{Variational Auto-encoders (VAEs)}: These methods\cite{kingma2013auto, doersch2016tutorial} work by transforming input samples into a latent embedding space, imposing a distribution on that space by adding noise as well as an appropriate KL divergence in the loss function, and then decoding the latent representations back into the original samples (using for example a Gaussian model). In order to use this as a generative model, one can sample and decode points from the latent space according to the imposed distribution. If the actual topology of the data distribution is poorly matched to the imposed distribution in the latent space, this method may assign a high likelihood to samples that are actually very unlikely. As such, there are a variety of methods which impose more flexible distributions on the latent space\cite{makhzani2015adversarial, dilokthanakul2016deep, jang2016categorical}. Additionally, by doing a series of Monte-Carlo-like resamplings of generated samples (essentially repeatedly encoding and decoding them), points which lie far from the data distribution end up getting mapped back into it\cite{creswell2016improving}.

VAEs are a very flexible framework --- multiple hierarchical latent spaces can be readily used, just by constructing the appropriate noise sources and KL divergence terms to add to the loss. One limitation is that one must specify a metric or parameterized probability distribution on the sample space --- generally mean-squared error, which corresponds to an isotropic Gaussian model. If this metric is poorly matched to the type of data being reconstructed, the resulting model tends to have a strong bias towards mode-collapsed or 'blurry' reconstructions,  essentially outputting the mean of the data distribution.

\textbf{Generative adversarial networks (GANs)}: Generative adversarial networks use a pair of networks in order to effectively estimate, then minimize, the difference between two data distributions. The discriminator network tries to classify samples as belonging to either the true or fake distribution, while the generator network uses the gradients propagated through the discriminator to try to fool it. The Nash equilibrium of this game is the point at which the fake and true distributions are identical. Due to training instability and phenomena such as mode collapse, there have been a number of refinements on the basic GAN idea such as LSGAN\cite{mao2017least}, Unrolled GANs\cite{metz2016unrolled}, WGAN\cite{arjovsky2017wasserstein}, WGAN-GP\cite{gulrajani2017improved}, Relativistic GANs\cite{jolicoeur2018relativistic}, as well as fusing GANs with other techniques such as in VAEGAN\cite{larsen2015autoencoding}, BEGAN\cite{berthelot2017began}, etc. Very high resolution outputs have been achieved by progressively stacking generators at a sequence of scales (Progressive GANs\cite{karras2017progressive}).

GANs can produce high-quality samples and retain the advantage of having a latent space to index the output distribution. However, training tends to be unstable, and among the various methods there tends to be a tradeoff between convergence and stability (with WGAN producing more stable training, but ultimately slightly worse converged results than the initial DCGAN\cite{radford2015unsupervised} formulation). GANs also do not produce a probability density associated with each sample. 

\section{Method}

Let us consider a problem in which a model is provided some information $x$ and is being optimized to minimize the expectation of a loss function comparing outputs to targets $E_{p(x,y)}\left[ L(\hat{y}(x),y) \right]$. From the information about $y$ contained in $x$, there is some conditional distribution $p(y|x)$ which could in principle be discovered e.g. in the limit of infinite data. What should the model output in this case?

If we consider the case of a particular value of $x$, then there is some distribution $p(y|x)$ of targets and therefore, for any single value of $\hat{y}$, some expectation over that distribution of the observed loss $\bar{L}(\hat{y},x) = E_{p(y|x)}\left[L(\hat{y},y)\right]$. Therefore, if the model produces some distribution of outputs $p(\hat{y}|x)$, the expected loss is:

\begin{equation}
E\left[L\right] = \int p(\hat{y}|x) \bar{L}(\hat{y},x) d\hat{y}
\end{equation}

This expectation is minimized when $p(\hat{y}|x)$ is a $\delta$-function at the global minimum value of $\bar{L}(\hat{y},x)$, and so the optimal behavior for a model when trained against the expectation of a loss function is to output a single value, even if spare entropy e.g. in the form of auxiliary noise inputs is available.

Now let us instead consider the case where, for each fixed $x$, we optimize the reduce to minimum over a set of loss values associated with different samples from some auxiliary noise distribution $p(\eta)$. This auxiliary distribution is added as an additional input to the network, meaning that $\hat{y}(x)$ becomes $\hat{y}(x,\eta)$ for each value sampled from $p(\eta)$. This gives rise to a new effective loss: process is:

\begin{equation}
L_{\textrm{EVL}} \equiv \min_\eta L(\hat{y}(x, \eta),y)
\end{equation}

So long as $\hat{y}(x,\eta) = y$ for some value of $\eta$, then this loss will be at a global minimum --- that is to say, this loss is minimized so long as $\hat{y}(x,\eta)$ covers the support of $y(x)$. If we approximate this loss with a finite but large number of samples from $p(\eta)$, then the loss can be reduced if the support of $\hat{y}(x,\eta)$ exactly matches the support of $y(x)$ rather than simply covering it. That is to say, the optimal strategy of the model in practice will be to align the support of the two distributions. 

Note that the actual probability densities $p(\hat{y})$ and $p(y)$ need not match in either of these cases to be at a global optimum --- it will instead converge to some related distribution based on the shape of the loss function around its optimum. If the loss function is concave up (e.g. it's behavior around zero scales as $(y-\hat{y})^q$ for $q>1$) then small deviations are weighted less significantly than large deviations, and so even for a small number of samples the tendency will be to rapidly produce a uniform distribution over the support. On the other hand, if the function is concave down ($q<1$) then small deviations are infinitely more significant than large deviations as the deviations approach zero, meaning that the model will tend to concentrate it's predictions according $p(y|x)$.

If we consider the other extremal loss, taking the maximum over the batch, then this reinforces the $\delta$-function output behavior even more strongly than the expectation would. 

\section{Connection between extreme value loss and mixture density networks}

An alternate method for modelling distributions is to parameterize a functional representation of the distribution itself, and then optimize the prediction of those parameters with respect to the log likelihood of the data under the corresponding predicted distribution. While any analytically tractable parameterization of a distribution may be used here, one convenient choice is to make use of a sum of $N$ Gaussian distributions, such that (in 1d, for example):

\begin{equation}
p(x, \vec{\rho}, \vec{\mu}, \vec{\sigma}) = \sum_i \rho_i (\sigma_i \sqrt{\pi})^{-1} \exp \left(-\frac{(x-\mu_i)^2}{\sigma_i^2}\right)
\end{equation}

with $\sum_i \rho_i = 1$.

If we consider the $N$ different Gaussian distributions to effectively be $N$ separate guesses, this method can be related to the idea of using a different function than the mean to summarize the losses over a set of samples. Although in a mixture density framework these guesses could be dependent on one-another, we can imagine the case in which the guesses were independently sampled from a distribution $\rho(\mu, \sigma)$. In that case, we can replace the expectation over a particular set of guesses $\sum_i \rho_i f(x,\mu_i, \sigma_i)$ with an expectation over the parameters which identify the distribution of guesses:

\begin{equation}
E_{\mu,\sigma} [ f(x, \mu, \sigma) ] = \int d\mu d\sigma \rho(\mu, \sigma) f(x, \mu, \sigma)
\end{equation}

For simplicity, we look at the case where $\sigma = 1$. In that case (up to a constant offset), the loss $L$ corresponding to the log-likelihood of the data given the model is:

\begin{equation}
L = E_x \left[ -\log p(x,\rho) \right] = E_x \left[ -\log E_\mu [ \exp(-(x-\mu)^2) ] \right]
\end{equation}

or, in integral form:

\begin{equation}
L = - \int dx p(x) \log \int d\mu \rho(\mu) \exp(-(x-\mu)^2)
\end{equation}

Note that, unlike a pointwise loss, the expectations do not commute with eachother because of the $\log$. That is to say, we cannot simply move the $x$ integral to inside the $\log$, find some effective loss, and then take the expectation over $\mu$ with respect to that effective function of $\mu$ alone. 

If we look at the innermost integral, we can approximate the integral by its value at the maximum of the function $\rho(\mu) \exp(-(x-\mu)^2)$:

\begin{equation}
L \approx \int dx p(x) \left( (x-\mu_*(x))^2 - \log \rho(\mu_*(x)) \right)
\end{equation}

If the $\rho$ term were constant, then this corresponds exactly to the mean-squared error loss evaluated at the closest-matching guess $\mu_*$ contained within the support of the distribution $\rho(\mu)$. With the $\rho$ term, however, there is an additional pressure that the distribution should be concentrated around the guesses $\mu_*$ which turn out to be close to $x$ --- which, practically speaking, would be the difference between optimizing $\rho(\mu)$ to be uniform over the support of $p(x)$, versus optimizing $\rho(\mu)$ to minimize divergence with $p(x)$.

\subsection{Generating the distribution}

Using the connection to mixture density networks, we can see that holding $\rho$ constant corresponds to modelling the support of the distribution, while allowing $\rho$ to vary to maximize the likelihood of the accepted guesses under a probabilistic model corresponds to modelling the distribution itself. We can obtain this same effect for the extremal loss method by having the network not just output coordinates of samples, but also training it to output an estimated probability for each sample of that particular sample being chosen as the extreme value --- corresponding to learning the function $\rho(\mu_*)$. This is done by outputting an extra channel which is then interpreted as the logits over the batch for a softmax function, then training against the categorical cross-entropy loss.

We then have two distributions: $q(\mu_*)$, corresponding to the pattern of samples produced by a push-forward transform of noise inputs to the network, and $\rho(\mu_*)$, which is a learned relative probability associated with each sample in a generated batch of samples. The distribution $q(\mu_*)$ is pushed towards capturing the support or the overall topology of the data, whereas the function $\rho(\mu_*)$ is pushed towards capturing variations in relative probability density among things which can be assumed to be within the support. The advantage of this method is that by factoring out explicit modelling of the likelihoods associated with points far away from the data manifold, the overall variation of the function $\rho(\mu_*)$ is reduced and it's behavior far from the boundaries of the data distribution becomes irrelevant. On the other hand, learning $\rho(\mu_*)$ directly means that the behavior of the function far from data can contribute strongly to the overall normalization factor while being only indirectly supervised from the data.

In order to generate samples to match the underlying distribution, we then simply generate a batch of samples according to $q(\mu_*)$ and reject samples based on the probability distribution over the batch given by $\rho(\mu_*)$: $p(\mu_*) = q(\mu_*) \rho(\mu_*)$.

\section{Low-dimensional Test Cases}

We first look at modelling distributions in sufficiently low dimension such that we may directly estimate divergences to the known true distribution via accumulating histograms over samples. This allows us to evaluate how well each method converges to the underlying distribution without relying on qualitative or indirect methods. We will consider both data formed of multiple Gaussians in up to 4 dimensions, as well as the 3d 'swiss roll' function.

Because the extremal loss method is expected to converge to the support of the distribution rather than the distribution itself, we use rejection sampling according to the probability predicted by the model that each sample from a batch will be chosen as the minimum. The network used in these experiments is a dense network comprised of 5 layers with ReLU activation functions, each of width 256, where the input is 16 dimensional unit-Gaussian noise. Each layer is initialized orthogonally\cite{saxe2013exact}, using $\sqrt{2}$ gain as appropriate to ReLU. The batch size is taken to be $200$, and for each sample from the batch we generate $128$ independent attempts over which we will compute the minimal (mean-squared error) loss. The network is trained over $50$ epochs using RMSprop\cite{tieleman2012rmsprop}, with a learning rate starting at $5\times10^{-4}$ and decreasing by a factor of $0.95$ each epoch. The mean-squared reconstruction loss and probabilistic loss are weighted equally.

These parameters were found via search against a validation set (corresponding to random seed 81379). The most significant contributors to performance were using RMSprop instead of Adam (which was introducing some oscillatory instabilities) and increasing the number of samples over which the extremal loss was evaluated. The decaying learning rate was important in obtaining convergence when close to the true distribution, though is not necessary for stability or for initial capture of the broad shape of the distribution.

We compare the extreme value loss network with three other baselines: an empirical estimate of the distribution, a Gaussian mixture model, and an unrolled relativistic GAN.

\textbf{Empirical distribution}: The simplest baseline is to estimate the empirical distribution via computing the histogram of the training data directly. In essence, this corresponds to approximating the distribution with a basis set of high-dimensional step functions. As a baseline, this has the advantage of strong convergence (given an infinite number of samples for both the training and test sets, the divergence will asymptotically be zero). However, as the dimensionality of the system increases, the empirical distribution is expected to become increasingly sample inefficient.

\textbf{Gaussian Mixture Model}: We use the Scikit-Learn\cite{pedregosa2011scikit} implementation of Gaussian mixtures, and set the number of components to predict to $10$ in all tests. As one of our test cases will be a data distribution which is actually sampled from a mixture of Gaussians, this baseline will let us determine how well a model can do when it has apriori correct knowledge of the (parameterized) function family from which the data is drawn, and when we compare to non-Gaussian datasets we can use this baseline to see how well those assumptions generalize.

\textbf{Unrolled relativistic GAN}: Our two baselines so far do not generalize well to high-dimensional cases such as the space of natural images. Generative adversarial networks, while being known for suffering from mode collapse issues, do successfully capture the characteristics of photorealistic images in the high-dimensional case. As such, we would like to compare with a GAN-based method. The basic GAN proved too unstable to produce consistent estimates of divergences to the data distribution, especially in the multi-modal cases. However, with a combination of unrolled GAN and relativistic GAN we were able to suppress oscillations and mode collapse, and observe reasonable convergence on a 4-Gaussian validation set in 2d (seed 81379). We used a network architecture similar in terms of depth, activation functions, and training method to that of the extreme value loss network for both the generator and discriminator, aside from the input and output dimensions, which are given by the GAN architecture constraints. Experiments with network geometry showed that a network width of $32$ was more stable than the value of 256 used for the extreme value loss network, and so we report results using that value.

\subsection{Evaluation}

We use two measures to evaluate the difference between the model distribution $p$ and test distribution $q$: the KL divergence and the Fisher information metric.

The KL divergence from the model distribution to the test distribution is:

\begin{equation}
KL(p||q) = -\sum_i p_i \log(q_i / p_i)
\end{equation}

In practice, if the supports of the empirical distributions disagree then this sum may diverge (either due to the argument of the $\log$ becoming zero due to $q_i=0$, or becoming infinite when $p_i = 0$). As such, we regularize the empirical distributions $p_i$ and $q_i$ by adding a virtual sample with weight $10^{-32}$ to each bin. When the supports disagree due to small-number statistics, each such bin contributes an error in estimating the KL divergence of order $\log 10^{-32} / M$ where $M$ is the total number of samples. This corresponds to an error floor of about $1.8 \times 10^{-4}$. When there is a systematic difference in supports, however, the way that this is weighted relative to differences in the probability densities depends systematically on the choice of this regularization term, and so the KL divergence becomes unreliable in that case.

To resolve this, we also look at the Fisher information metric between the distributions, which can be measured via the geometric mean of the distributions\cite{itoh2017geometric}:

\begin{equation}
F = 2\cos^{-1} \left( 1 - \sum \sqrt{p_i q_i} \right)
\end{equation}

This metric does not diverge if the supports disagree, and so is more stable against sampling noise in estimating the empirical distribution.

\subsection{Results}

\subsubsection{Gaussian Dataset}
We use data sampled from mixtures of $N$ unit-Gaussians of equal magnitude in 1d, 2d, 3d, and 4d. We look at datasets comprised of 1, 2, 4, and 10 Gaussians in order to investigate the degree to which the methods can capture multi-model distributions. To produce these datasets, we generate unit-standard deviation Gaussians whose centers are uniformly positioned over the range $[-6,6]$. This range is selected so that even in 1d we will tend to see multiple distinct peaks as we increase the number of modes. For each dimension and number of Gaussians, we generate 5 datasets (random seeds from $1$ to $5$) and train each of the baseline models and the extreme value loss network.

\begin{figure*}
 \includegraphics[width=\textwidth]{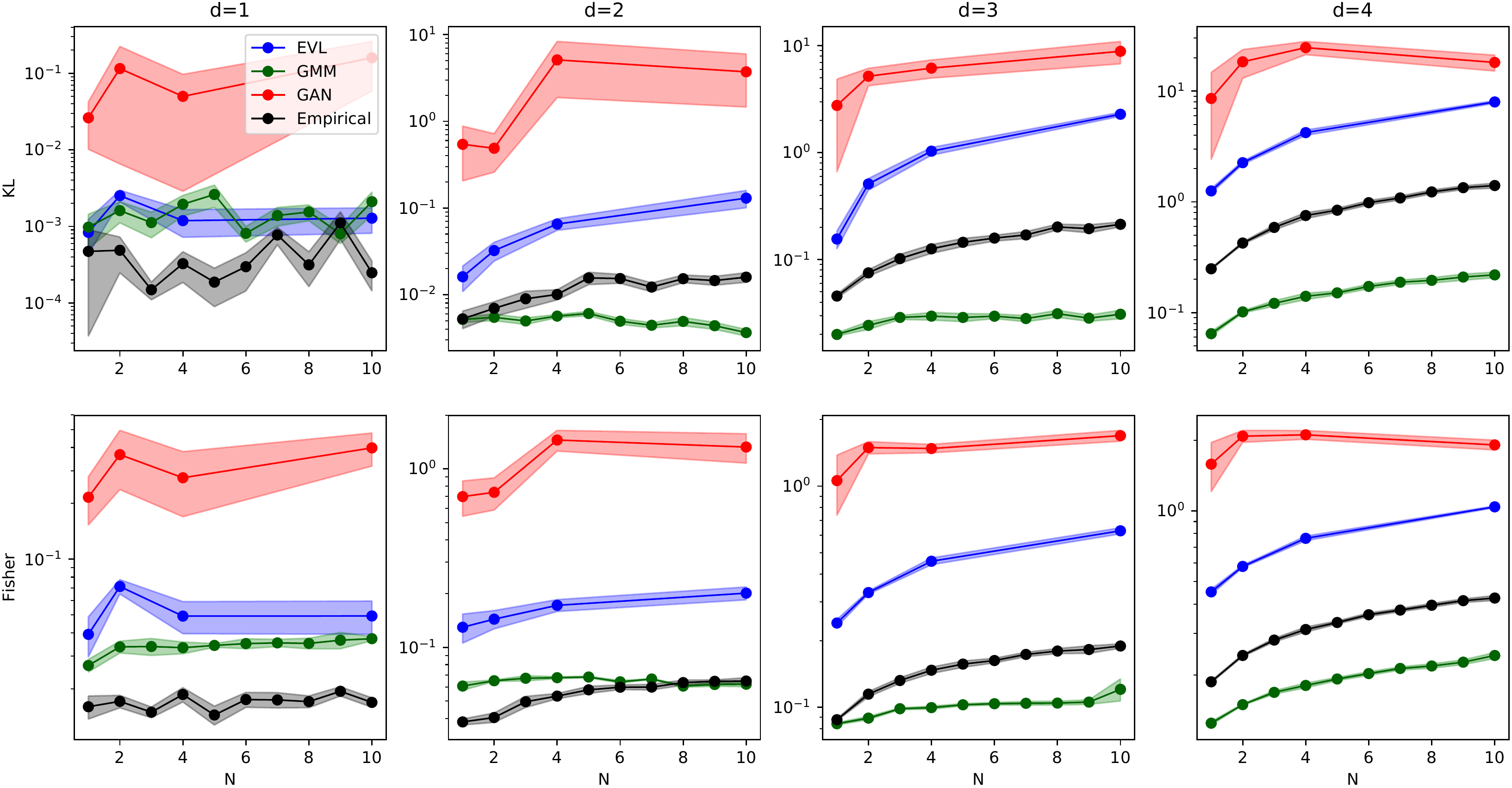}
 \caption{\label{KL_and_Fisher}KL divergence and Fisher metric between models and the underlying generated test dataset of $4\times10^{5}$ points for multiple Gaussians in different numbers of dimensions and numbers of Gaussians. Results for a training set size of $5\times10^4$. }
\end{figure*}

\begin{figure*}
 \includegraphics[width=\textwidth]{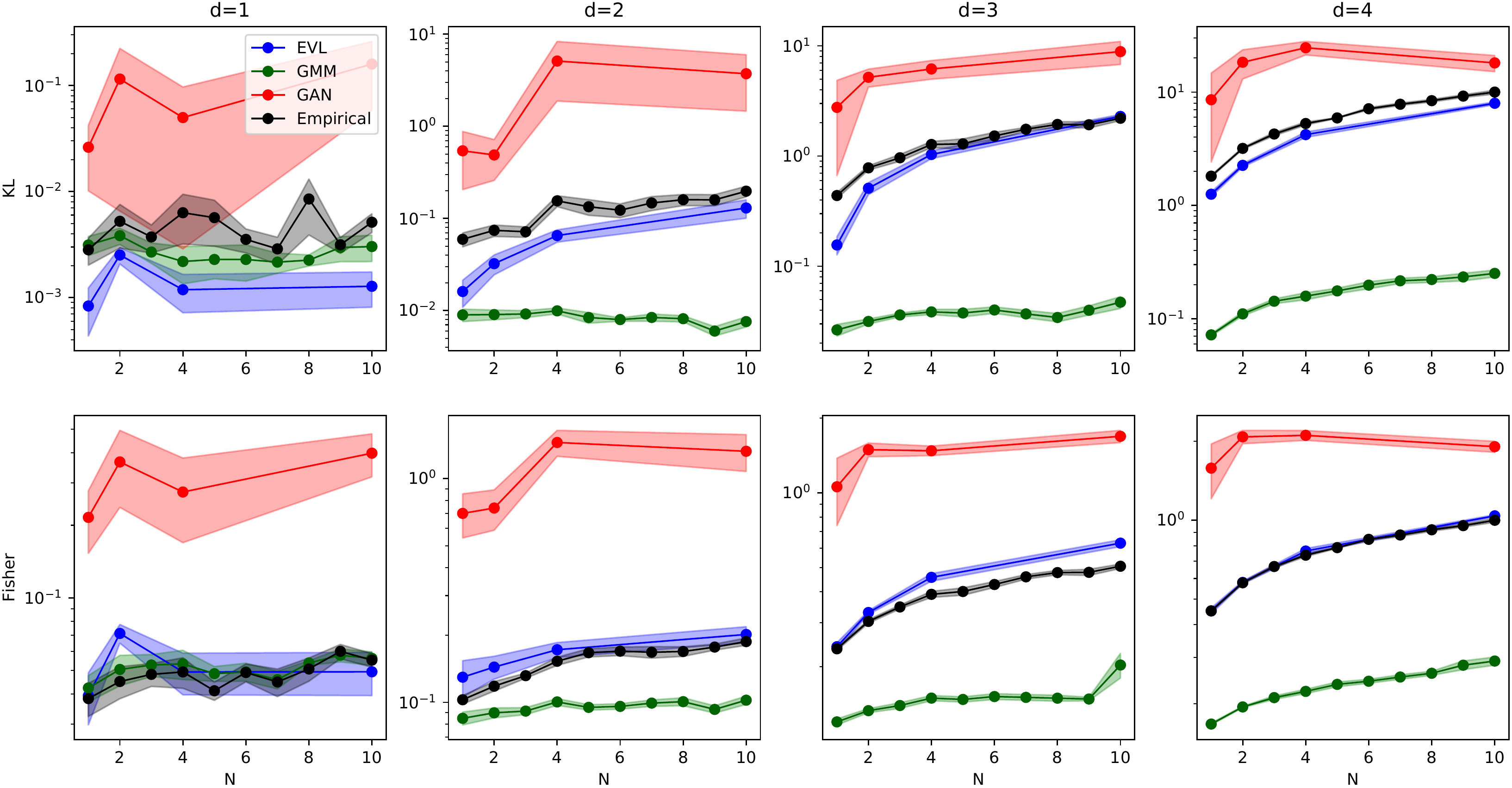}
 \caption{\label{KL_and_Fisher_5k}KL divergence and Fisher metric between models and the underlying generated test dataset of $4\times10^{5}$ points for multiple Gaussians in different numbers of dimensions and numbers of Gaussians. Results for a training set size of $5\times10^3$.}
\end{figure*}
 
The results for a training set size of $5\times10^4$ are shown for the KL divergence and Fisher metric in Fig.~\ref{KL_and_Fisher}. In general, the ordering of the four models is almost universally preserved across this set of experiments, with the Gaussian Mixture Model (GMM) performing the best, the empirical distribution second, the extreme value loss network (EVL) third, and finally the unrolled relativistic GAN. The GMM incorporates correct ground-truth information as to the nature of the underlying distribution, so it's performance should not be considered particularly surprising. When it comes to the other models, for this number of samples at least we can see that while the GAN and EVL do attempt to capture the structure of the empirical distribution (which comprises the training set), they do so incompletely. We note that along with having a higher divergence from the underlying data distribution, the GAN training also results in significantly larger error bars --- the underlying distribution of errors between runs is not Gaussian, but rather appears to be bimodal, suggesting that this may just indicate that the GAN occasionally fails to converge entirely, but may perform better than indicated by the mean in cases in which it does not totally diverge.

While the empirical distribution outperforms the two network-based approaches here, if we look at the absolute scale of the divergences we note that they are quite small in 1d and 2d: about $10^{-3}$ nats in 1d and $10^{-1}$ nats in 2d as the worst cases. As such, if the purpose for training the model is not just to generate samples but, for example, to learn a differentiable parameterization (which would not be available from either the empirical baseline or Gaussian Mixture baseline) then this may be sufficiently good to pursue that end. 

We also consider what happens in the limit of smaller training sets comprised of $5000$ samples. The results are shown in Fig.~\ref{KL_and_Fisher_5k}. In this case, the empirical distribution becomes extremely sparse in the higher dimensional cases. Here we can see that the extreme value loss network actually outperforms the empirical distribution (despite having no explicit prior knowledge about the function space) in terms of KL divergence, and performs comparably with respect to the Fisher metric. This difference makes sense in the context of the extreme value loss having a bias towards ensuring overlapping support, which is more important with respect to the KL divergence than it is to the Fisher metric.

\subsubsection{Swiss roll dataset}

\begin{figure*}
 \includegraphics[width=\textwidth]{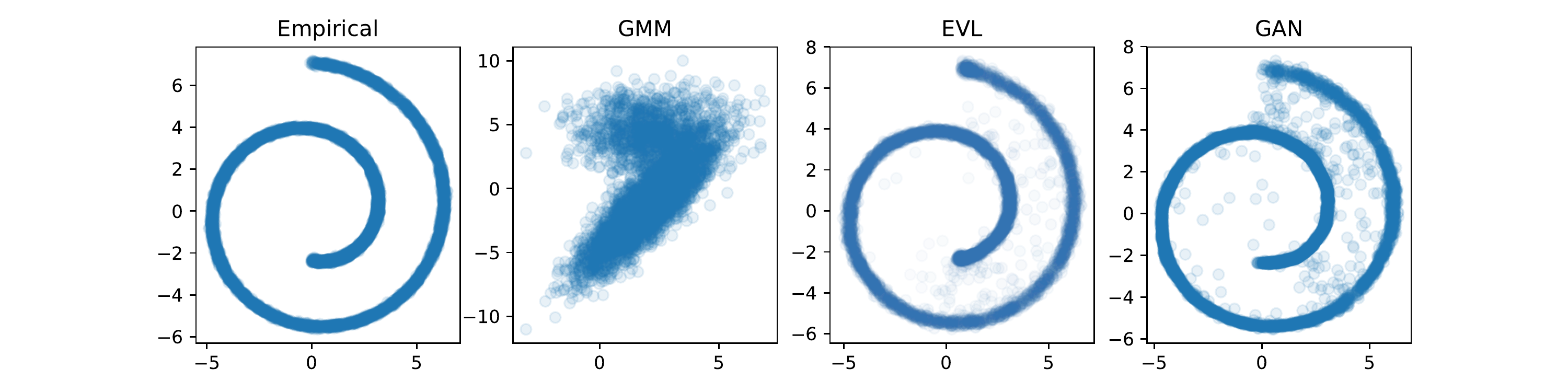}
 \caption{\label{Swissroll} Learned swiss roll distributions from the various models. a) Empirical distribution (training set), b) Gaussian Mixture Model, c) Extreme Value Network, d) Unrolled GAN}
\end{figure*}

Here we present results of the extreme value loss on the swiss roll dataset (as generated by Scikit-Learn's \textbf{make\_swiss\_roll} function with noise set to $0.1$, and scaled by $0.5$ so that they line within the same domain as the multi-Gaussian test case presented above). We generate $5\times10^4$ training set values and $4\times10^5$ test set values, and compare the various methods discussed in the prior section. The same parameters as in the previous section are used for all methods. The resulting generated distributions are shown in Fig.~\ref{Swissroll}, projected down into the 2 varying dimensions (the third dimension of the swiss roll dataset is included for evaluating divergences, and has significant consequence for the performance of the Gaussian Mixture Model in particular). The corresponding KL divergence and Fisher metric for the various methods are reported in  Table~\ref{SwissrollValues}.

\begin{table}
\caption{Comparison of methods on swiss roll dataset}
\label{SwissrollValues}
\centering
\begin{tabular}{|c|c|c|}
\hline
Method & KL & Fisher \\
\hline
Empirical & 0.054 & 0.126 \\
GMM & 0.312 & 1.294 \\
Extreme Value Loss & 0.204 & 0.411 \\
Unrolled GAN & 0.246 & 0.647 \\
\hline
\end{tabular}
\end{table}

Notably, while the empirical distribution is still the best representation of the underlying distribution for this amount of data, the Gaussian Mixture Model now fails almost completely to capture the structure of the dataset (which makes sense, as the underlying function spaces now differ). On the other hand, the unrolled relativistic GAN and the extreme value loss network both manage to capture the structure of the dataset and sharply reproduce the spiral pattern, though the extreme value loss network obtains quantitatively better KL divergence and Fisher metric values.

\section{High-dimensional data}

In the previous case, we consider the extreme value loss in the context of low dimensional data distributions. However, the usual context in which neural network based generative modelling is considered is in the production of realistic high-dimensional objects such as images or sounds. Here we hit a potential limitation of the extreme value loss approach to generating diverse distributions. Each time the network generates a set of guesses, those guesses should be distributed such that there is a high probability that at least one guess is close to the target point. As such, the guesses should be positioned to cover the space of fluctuations of the target. If the target point has fluctuations which are inherently high-dimensional (with respect to the choice of metric used to compare points), then the number of guesses needed to cover that space grows exponentially with the dimension. Correspondingly, the benefit to be gained by adding additional guesses will be exponentially reduced compared to simply making a single guess. On the other hand, if the target point lies on a low-dimensional manifold, then this may not pose a problem and the network will learn a map over that subspace of fluctuations.

We consider two image datasets --- MNIST\cite{lecun2010mnist} digits, and CIFAR10\cite{krizhevsky2014cifar}. The MNIST dataset has objects of dimensions $28 \times 28$, while CIFAR10 has objects of dimension $32 \times 32 \times 3$, so these are roughly comparable (and both are much higher in dimension than the cases we considered in the previous section). However, while MNIST digits are fairly regular (to the extent that the image of the 'average digit' has recognizable structures in it), the CIFAR images are much higher entropy due to factors such as variations in the background and lighting.

It is well-established that carefully calibrated generative adversarial networks can produce realistic samples from image datasets, whether or not they model the actual underlying distribution of images. Datasets such as MNIST are easy problems in this context, with modern GANs (such as Progressive GANs) being able to attain resolutions up to $1024^2$. As such, rather than training an MNIST or CIFAR GAN baseline to compare with, we keep in mind that from the point of view of alternate methods this is essentially a solved problem. Our interest, in this section, is more to understand the failure modes of using a method such as extreme value loss to train generative models as the dimensionality of the output becomes large.

In the following experiments, we tried a number of variations which made little difference, and so some hyperparameter choices are kept for ease of visualization rather than performance. Specifically, we only use a 2d latent space of input noise (increasing this to 16d led to no appreciable improvement on either MNIST or CIFAR sample quality). This allows us to render flat maps of the network's entire generated space, which helps directly evaluate things such as the relative probability of each MNIST digit, the relationship between features, and the characteristics of interpolations across the space by eye. On the other hand, we found that the width of the network was quite important in terms of obtaining sharp samples in the case of MNIST, and increasing the network dimensions also led to some improvements in CIFAR sample quality. The final size of the networks we use here was chosen based on the 8gb GPU memory limit of our hardware rather than a measure of sample quality, and it seems as though increasing the network sizes further might lead to further (small) improvements. Increasing the number of guesses also led to small improvements in sample sharpness and quality, particularly in the case of the CIFAR images, but was also constrained by memory considerations.

The details of the architecture are as follows. We use a convolutional neural network architecture composed of blocks, where each block contains a linear upscaling operation that doubles the current image resolution, followed by two convolutional layers with $3 \times 3$ kernels and ReLU activations. The number of filters in each block are 512, 256, 128, 128, and 64 respectively. A final linear $3\times3$ convolution converts the output of this stack into the appropriate pixel representation (1 channel for MNIST, 3 channels for CIFAR). In parallel, the input noise vector is passed through 4 dense layers comprised of 256 units each and ReLU activations, and a final projection down to 1d which is then passed through a softmax activation to obtain the probability estimate that each guess is selected as the closest one (so that we can reconstruct the distribution through rejection sampling). All layers are orthogonally initialized, with a gain of $\sqrt{2}$ in the case of ReLU activations. We use RMSprop to optimize the model, with an initial learning rate of $10^{-4}$ and an exponential decay by a factor of $0.95$ each epoch, and train the model for $20$ epochs. The loss function to compare generated samples to the target is mean squared error (we expect that to get photorealistic samples, this would need to be replaced with a perceptual loss or adversarial loss).

Because this task is more memory intensive than the previous ones, we make use of the fact that the network's output is independent of the sample that it is trying to match, and so rather than generating independent random samples for each image from the batch, we match each image from the training batch to its best match from a fixed generated set per batch. That is to say, if we use a batch size of 200 images and 128 generated attempts, rather than generating $200 \times 128$ images we only need to generate $128$, which we can then match. This allows us to scale the approach to larger sampling rates, which increases the probability of finding significant fluctuations. As a result, we are able to use a batch of size $50$ and generate $512$ guesses per pass (these guesses use the same noise vectors over a single batch, but independent ones between batches).

\begin{figure*}
 \includegraphics[width=\textwidth]{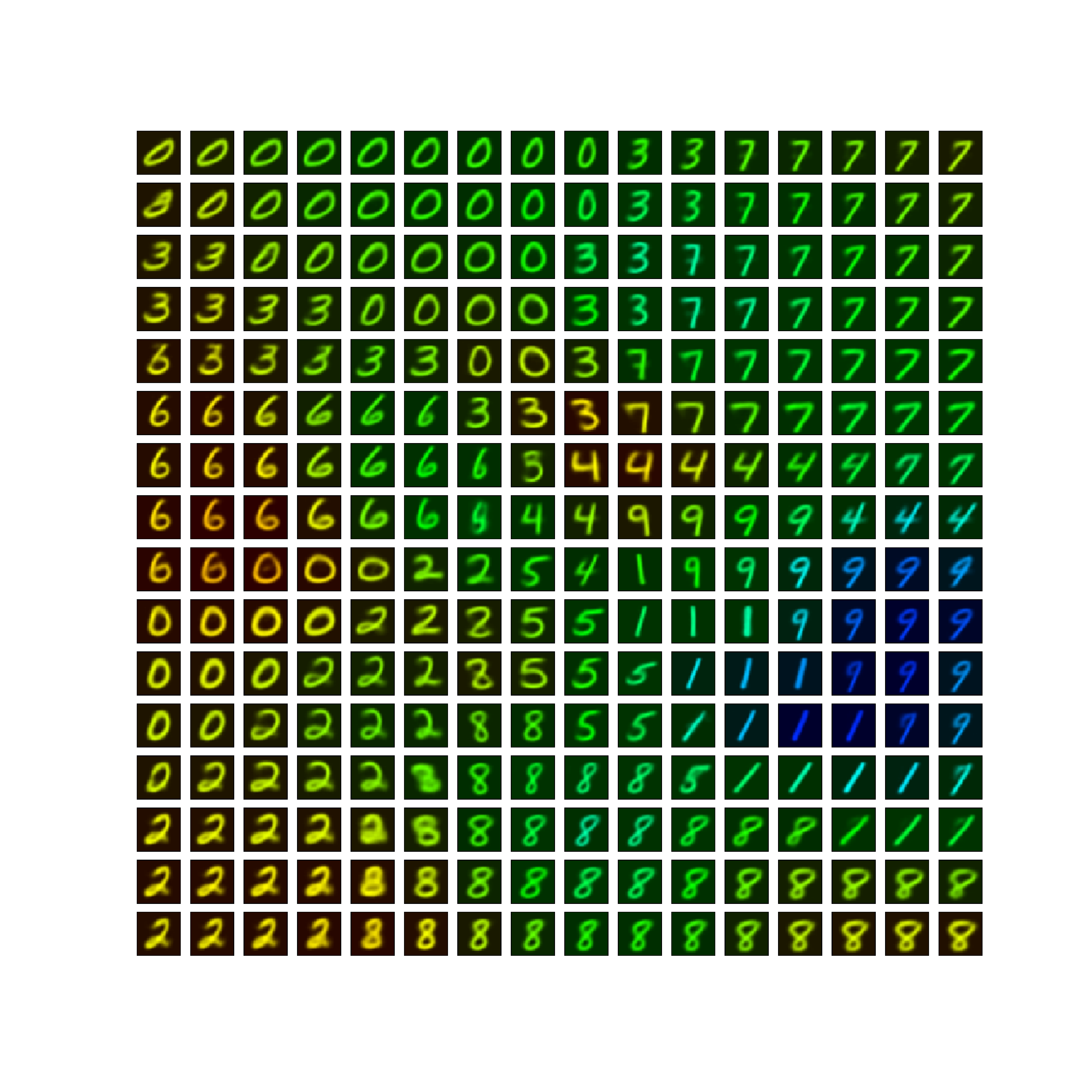}
 \caption{\label{MNIST}Generated MNIST digits arranged according to the 2d latent code which generates them. The color of each image shows the relative probability given by the network of that sample being selected as the closest match to a random example from MNIST, with blue corresponding to high probability and red corresponding to low probability.}
\end{figure*}

\begin{figure*}
 \includegraphics[width=\textwidth]{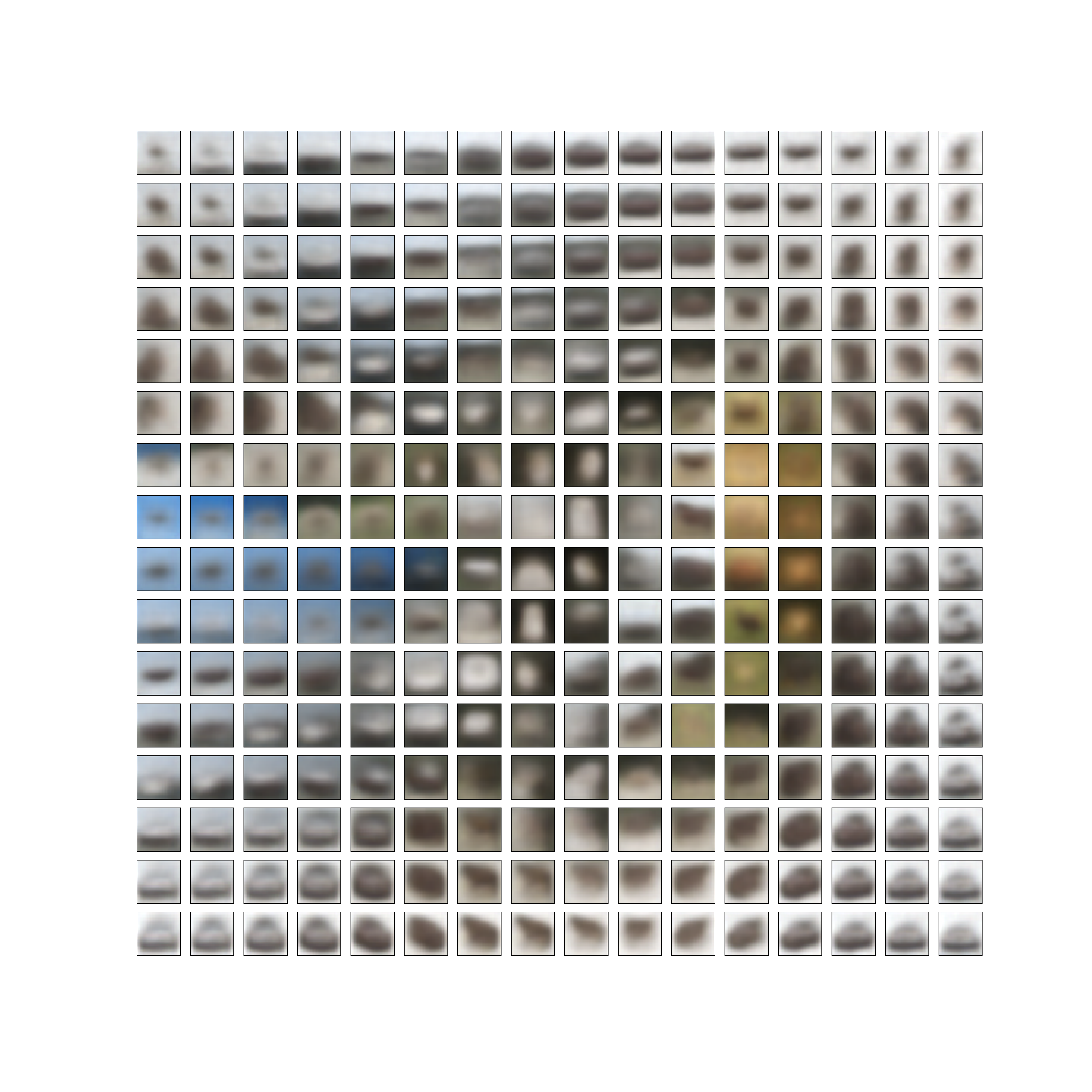}
 \caption{\label{CIFAR}Generated CIFAR-10 samples, arranged according to their 2d latent code. No fine details are visible, despite this architecture being able to handle similar resolution outputs in the case of MNIST. Instead, the 2d latent space is mapped according to broad features such as overall color, gradients, and the positioning and shading of a generic central shape.}
\end{figure*}

The results on MNIST and CIFAR are shown in Fig.~\ref{MNIST} and Fig.~\ref{CIFAR} respectively. While we see relatively sharp reconstructions of the MNIST digits, it seems clear that while the CIFAR examples have reconstructed variations in the background and gradient, the pixel-level variations have been blurred out. This may in part be because of the mean squared error, which focuses on pixel-level correspondences and as such reconstructs long-wavelength features in preference over short-wavelength features. If, for example, we used Gram matrices on VGG activations, we might expect to see a map covering the distribution of visually distinct textures instead. However, in either case, the entropy of the CIFAR images means that the network is not able to latch onto representative examples as it can do with MNIST.

In practice, this means that while extreme value loss networks on their own can stably match parts of distributions, in order to scale the technique to high dimensional data we would need a better way of associating a given target point with a closely matching guess (and, correspondingly, to correctly track the probability density associated with that guess). This may involve approaches such as an internal gradient descent step, factorization of the noise latents into a series of sequential decisions, or other techniques. Our purpose in this paper is to introduce the basic idea and map out it's strengths and weaknesses towards different kinds of data and problems, so we leave these elaborations to future work.

\section{Conclusions}

We observed that the default tendency of training against the expectation of a loss function over a dataset is to converge to outputting $\delta$-function distributions around a point estimate related to the conditional distribution, even when those point estimates are not representative of the distribution as a whole. However, by allowing the model to make multiple guesses and then selecting only the best guess to contribute to the overall loss, the model is rewarded for spanning the support of the distribution of possible targets. In combination with a model that predicts which guess will be chosen, this may be used to train a generative model that reproduces the distribution of targets, provides relative probability density estimates in that space, and learns a continuous map from a latent space into a corresponding manifold of representative targets which span the data.

We demonstrate that the technique converges to the distribution in low-dimensional cases in which we can directly estimate the KL divergence and Fisher Information Metric between generated and target distributions, and that the pattern of training is generally robust and reliably convergent. We also investigated the limits of the method, and showed that while high dimensional outputs do not necessarily pose a problem, when the outputs lie on a high-dimensional or high-entropy manifold then that can pose a problem and prevent the model from learning details of the data. We suggest that this may be alleviated by improving the quality of the random guesses through a variety of inner-optimization-loop strategies, essentially refining the guess so that fewer overall samples are needed in order for the model to take advantage of a large potential expressive space.

The practical applications of this technique most likely involve handling systems in which there is some highly non-differentiable or stochastic element of the real map from input to output --- for example, chaotic dynamical systems --- but where one would still wish to for example take derivatives of summary statistics over the output relative to inputs or parameters. By indexing the distribution of possibilities, networks trained in this way learn representations which summarize the distribution rather than representing individual points. As such, while a chaotic system may exhibit a large change in outcome in response to a small change in initial conditions, it will not necessarily exhibit a large change in \textit{statistics} in response to a small change in initial conditions, driving forces, or other parameters. The extreme value loss approach enables networks trained on regression tasks to capture that distinction, and as such may result in more stable long-term dynamics predictions, as well as a better ability to use learned models for differentiable planning.

Another potential application arises not from the stability and generative nature of models trained this way, but from the observation that the model learns to generate the support rather than the actual distribution. In importance sampling, one wishes to re-weight their sampling strategy in order to reduce the variance of the estimator of a target observable as quickly as possible. This is ideally done by weighting each sample by the inverse of the probability density of the corresponding value of the observable associated with that sample. In essence, one wants to sample from a uniform distribution over the possible values of the observable --- the supporting distribution --- in order to maximize the convergence rate of estimates of that observable. Since a network trained against an extreme value loss rather than expectation over the loss converges to the support rather than the distribution, sampling from the network could be used as part of an importance sampling scheme. 

\section{Acknowledgements}

We would like to acknowledge Martin Biehl for helpful discussions, and Brent Werness for the suggestion to use the Fisher metric to compare distributions with differing support.

\bibliography{evl}

\end{document}